\documentclass[]{interact}

\usepackage{epstopdf}
\usepackage[caption=false]{subfig}

\usepackage[numbers,sort&compress]{natbib}
\bibpunct[, ]{[}{]}{,}{n}{,}{,}
\renewcommand\bibfont{\fontsize{10}{12}\selectfont}
\makeatletter
\def\NAT@def@citea{\def\@citea{\NAT@separator}}
\makeatother

\theoremstyle{plain}

\theoremstyle{definition}

\theoremstyle{remark}

\usepackage{multirow}
\usepackage{booktabs}

\usepackage[labelsep=space,singlelinecheck=false]{caption} 
\usepackage{amssymb}
\usepackage{amsthm}
\usepackage{amsmath}
\usepackage{bm}

\newcommand{\bR}{\mathbb{R}}

\newcommand{\bx}{\bm{x}}

\newcommand{\cN}{\mathcal{N}}
\newcommand{\cD}{\mathcal{D}}

\usepackage{algorithm} 
\usepackage{algorithmicx}
\usepackage{algpseudocode} 
\algnewcommand\Input{\item[\textbf{Input:}]}
\algnewcommand\Output{\item[\textbf{Output:}]}

\usepackage{xcolor}

\begin{document}






\title{Approximate Gibbs Sampler for Efficient Inference of Hierarchical Bayesian Models for Grouped Count Data}

\author{
\name{Jin-Zhu Y\"u\textsuperscript{a,b,*}\thanks{*CONTACT Jin-Zhu Y\"u. Email: yujinzhu88@gmail.com. 
} and Hiba Baroud\textsuperscript{c,d}}
\affil{\textsuperscript{a}Department of Civil Engineering, University of Texas at Arlington, Arlington, TX, USA; 
\textsuperscript{b}Department of Industrial, Manufacturing, and Systems Engineering, University of Texas at Arlington, Arlington, TX, USA;
\textsuperscript{c}Department of Civil and Environmental Engineering, Vanderbilt University, Nashville, TN, USA;
\textsuperscript{d}Department of Computer Science, Vanderbilt University, Nashville, TN, USA}
}

\maketitle


\begin{abstract}
Hierarchical Bayesian Poisson regression models (HBPRMs) provide a flexible modeling approach of the relationship between predictors and count response variables. The applications of HBPRMs to large-scale datasets require efficient inference algorithms due to the high computational cost of inferring many model parameters based on random sampling. Although Markov Chain Monte Carlo (MCMC) algorithms have been widely used for Bayesian inference, sampling using this class of algorithms is time-consuming for applications with large-scale data and time-sensitive decision-making, partially due to the non-conjugacy of many models. To overcome this limitation, this research develops an approximate Gibbs sampler (AGS) to efficiently learn the HBPRMs while maintaining the inference accuracy. In the proposed sampler, the data likelihood is approximated with Gaussian distribution such that the conditional posterior of the coefficients has a closed-form solution. Numerical experiments using real and synthetic datasets with small and large counts demonstrate the superior performance of AGS in comparison to the state-of-the-art sampling algorithm, especially for large datasets.
\end{abstract}

\begin{keywords}
Conditional conjugacy; Approximate MCMC; Gaussian approximation; Intractable likelihood
\end{keywords}





\section{Introduction}\label{sec:intro}

Count data are frequently encountered in a wide range of applications, such as finance, epidemiology, sociology, and operations, among others~\cite{aktekin2018sequential}. For example, in epidemiological studies, the occurrences of a disease are often recorded as counts on a regular basis~\cite{hay2001bayesian}. Death counts, classified by various demographic variables, are regularly recorded by government agencies~\cite{de2013hierarchical}. In customer service centers, the service level is often measured based on the number of customers served during a given period of time. More recently, data-driven disaster management approaches have used count data to analyze the impact of disasters (e.g., number of power outages~\cite{han2009estimating} and pipe breaks~\cite{yu2021hierarchical}) and the recovery process (e.g., recovery rate~\cite{yu2019quantifying}). Understanding the features that can influence the occurrence of such events is critical to inform future decisions and policies. Therefore, statistical models have been developed to accommodate the complexity of count data, among which are Hierarchical Bayesian Poisson regression models (HBPRMs) that have been widely employed to analyze count data under uncertainty~\cite{ma2008multivariate,baio2010bayesian,flask2014segment,khazraee2018bayesian}. The wide applicability of this class of models is due to the fact that the hierarchical Bayesian approach offers the flexibility to capture the complex hierarchical structure of count data and predictors by estimating different parameters for different data groups, thereby improving the estimation accuracy of parameters for each group. The data can be grouped based on geographical areas, types of experiments in clinical studies, or different hazard types and intensities in disaster studies. The hierarchical structure assumes that the parameters of the prior distribution are uncertain and characterized by their own probability distribution with corresponding parameters referred to as hyperparameters. Therefore, this class of models can account for the individual- and group-level variations in estimating the parameters of interest and the uncertainty around the estimation of hyperparameters~\cite{gelman2013bayesian}.

The flexibility of hierarchical models in capturing the complex interactions in the data comes with a high computational expense since all the model parameters need to be estimated jointly~\cite{betancourt2015hamiltonian}. Furthermore, large-scale data may be structured in many levels or groups~\cite{aljadda2014pgmhd}, resulting in a large number of parameters to learn for a hierarchical model, further increasing the computational load. Given that many of the applications involving count data have recently benefited from technological advances in data collection and storage, there is a critical need to ensure the applicability of HBPRMs. As a result, efficient inference algorithms are needed to support the use of statistical learning models such as HBPRMs in risk-based decision-making, especially for time-sensitive applications such as resource allocation during emergency response and disaster recovery.

The most popular algorithms for parameter inference in hierarchical Bayesian models (and generally for Bayesian inference) are Markov Chain Monte Carlo (MCMC) algorithms. MCMC algorithms obtain samples from a target distribution by constructing a Markov chain (irreducible and aperiodic) in the parameter space that has precisely the target distribution as its stationary distribution~\cite{brooks1998markov}. This class of algorithms provide a powerful tool to obtain posterior samples and then estimate the parameters of interest when the exact full posterior distributions are only known up to a constant and direct sampling is not possible~\cite{brooks1998markov}. However, a major drawback of standard MCMC algorithms, such as the Metropolis-Hastings algorithm (MH), is that they suffer from slow mixing, requiring numerous Monte Carlo samples that grow with the dimension and complexity of the dataset~\cite{conrad2016accelerating,robert2018accelerating}. In some applications of Bayesian approaches (e.g., emergency response), decisions relying on outcomes of the model cannot afford to wait days for running MCMC chains to collect a sufficiently large number of posterior samples. As such, the application of standard MCMC algorithms to learn Bayesian models such as HBPRMs or other hierarchical Bayesian models for large datasets is significantly limited and a fast approximate MCMC is needed.

The key idea of approximate MCMC is to replace complex distributions that lead to a computational bottleneck with an approximation that is simpler or faster to sample from than the original~\cite{alquier2016noisy,johndrow2015optimal}. Several studies have applied analytical approximation techniques by exploiting conjugacy to accelerate MCMC-based inference in hierarchical Bayesian models~\cite{streftaris2008efficient,chan2009bayesian,dutta2016bayesian,berman2019asymptotic}. 
More specifically, an approximate Gibbs sampling algorithm to is used to enable the inference of the rate parameter in the hierarchical Poisson regression model in~\cite{streftaris2008efficient}. The conditional posterior of the rate parameter, which does not have a closed-form expression due to non-conjugacy between Poisson likelihood and log-normal prior distribution, is approximated as a mixture of Gaussian and Gamma distributions using the moment matching method. The exact conditional moments are obtained by minimizing the Kullback-Liebler divergence between the original and the approximate conditional posterior distributions. 
Conjugacy is also employed to improve inference efficiency in large and more complex hierarchical models in~\cite{dutta2016bayesian}. It is shown that the approximation using conjugacy can be utilized even though the original hierarchical model is not fully conjugate~\cite{dutta2016bayesian}. As an example in their study, the approximate full conditional distributions are derived when the likelihood function follows a gamma distribution while the prior for the parameters are assumed to be multivariate normal and inverse Wishart distribution. 
In~\cite{berman2019asymptotic}, a Gaussian approximation to the conditional distribution of the normal random effects in the hierarchical Bayesian binomial model (HBBM) is derived using Taylor series expansion, such that Gibbs sampling can be applied to infer the HBBM more efficiently. A similar approach that approximates the data likelihood with a Gaussian distribution to allow for faster inference of parameters is used for parameter inference in a Bayesian Poisson model~\cite{chan2009bayesian}. 
With regard to count data, a fast approximate Bayesian inference method is proposed to infer a negative binomial model (NB) in~\cite{bansal2020fast}. The non-conjugacy of the NB likelihood is addressed by the P\'{o}lya-Gamma data augmentation. This technique is first developed 
in~\cite{polson2013bayesian} and is employed to approximate the likelihood as a Gaussian distribution. Consequently, the conditional posteriors of all but one parameters have a closed-form solution and a Metropolis-within-Gibbs algorithm is thus developed for the posterior inference. 
 
While approximate MCMC algorithms have been developed for hierarchical and non-hierarchical Poisson models as well as hierarchical Bayesian binomial and negative binomial models, the development of approximate MCMC algorithm for an efficient inference of HBPRMs for grouped count data is still lacking. In this paper, we propose an approximate Gibbs sampler to address this problem. To deal with the non-conjugacy between the likelihood and the prior, we approximate the conditional likelihood as a Gaussian distribution, leading to closed-form conditional posteriors for all model parameters. The contribution lies in the derivation of a closed-form approximation to the complex conditional posterior of the parameters and the development of the Approximate Gibbs sampling (AGS) algorithm. The proposed algorithm allows for an efficient inference of the general HBPRM using the approximate Markov chain without compromising the inference accuracy, enabling the use of HBPRMs in applications with large-scale data and time sensitive decision-making. To demonstrate the performance of the proposed AGS algorithm, we conduct multiple numerical experiments and compare the inference accuracy and computational load to state-of-the-art sampling algorithms. Note that due to the use of Gaussian approximation, the AGS algorithm performs well when the dataset does not contain excessive zero counts.

The rest of this paper is organized as follows. In Sec. \ref{sec:meth}, a general hierarchical Bayesian Poisson model for grouped count data is presented, and the closed-form solution to the approximate conditional posterior distribution of each regression coefficients is derived, followed by a description of the proposed AGS algorithm. Sec. \ref{sec:experiments} introduces the datasets used in the numerical experiments along with the comparison of the performance of sampling algorithms. Conclusions and future work are provided in Sec. \ref{sec:conc}.



\section{Methodology}\label{sec:meth}

\subsection{Hierarchical Bayesian Poisson Regression Model}

This section presents the Hierarchical Bayesian Poisson Regression Model (HBPRM) for count data. Without loss of generality, we consider a general HBPRM, the hierarchical version of Poisson log-normal model \cite{streftaris2008efficient, aguero2009bayesian, montesinos2017bayesian, serhiyenko2016fast} for grouped count data, in which the coefficient for each covariate varies across groups (Eq. \eqref{eq:hbprm_first} to Eq. \eqref{eq:hbprm_last}). This model can be applied to count datasets in which the counts can be divided into multiple groups based on the covariates.
Let $\cD =\{x, y\}$ be the dataset where $x$ 
represents the covariates and $y$ represents the dependent positive counts. 
This HBPRM assumes that each count, $ y_{ij}$, follows a Poisson distribution. The log of the mean in the Poisson distribution is a linear function of the covariates. In the hierarchical Bayesian paradigm, each of the parameters (regression coefficients) in the linear function follows a prior distribution with hyperparameter(s) which are in turn specified by a hyperprior distribution. Note that the hyperpriors are shared among the parameters of the same covariate for all groups, thereby resulting in shrinkage of the parameters towards the group-mean and facilitating strength borrowing across groups \cite{gelman2013bayesian}. When the variance of the hyperprior is decreased to zero, the hierarchical model is reduced to a non-hierarchical model. The mathematical formulation of the HBPRM is provided in Eq. \eqref{eq:hbprm_first} to Eq. \eqref{eq:hbprm_last}:
\begin{align}
&     y_{ij}|     \lambda_{ij} \sim \textrm{Pois}(     \lambda_{ij}), \; \forall\; i=1,\dots,n_j,\; j=1,\dots,J,\label{eq:hbprm_first}\\
&\ln      \lambda_{ij} = \sum\limits_{k=1}^{K} {     w_{jk}      x_{ijk}}, \; \forall\; i=1,\dots,n_j,\; j=1,\dots,J,\;k=1,\dots,K ,\\
&     w_{jk}|\mu_{ k},\sigma_{ k}^2 \sim \cN (\mu_{ k},\sigma_{ k}^2),\; \forall \; j=1,\dots,J,\; k=1,\dots,K,\\
&\mu_{ k}|m,\tau^2 \sim \cN (m,\tau^2),\; \forall \; k=1,\dots,K,\\
&\sigma_{ k}^2|a,b \sim \mathcal{IG}(\frac{a}{2}, \frac{b}{2}), \; \forall \; k=1,\dots,K. \label{eq:hbprm_last}
\end{align}

\noindent In the HBPRM formulation, $ y_{ij}$ is the $i$-th count within group $j$ with an estimated mean of $     \lambda_{ij}$, $n_j$ is the number of data points in group $j$, $     w_{jk}$ is the regression coefficient of covariate $k$, and $     x_{ijk}$ is the $i$-th value in group $j$ of covariate $k$. 
The prior for the coefficient of each covariate, $\mu_{ k}$, is assumed to be a Gaussian distribution ($\cN$) while the prior for the variance, $\sigma^2_{ k}$, is assumed to be an inverse-gamma distribution ($\mathcal{IG}$). The Gaussian and inverse-gamma distributions are specified such that we can exploit conditional conjugacy for analytical and computational convenience. Alternative distributions (such as half-Cauchy and uniform distributions) for the prior of group-level variance $\sigma_k^2$ do not have this benefit, which will significantly increase the computational load. According to Ref. \cite{gelman2006prior}, when the group-level variance is close to zero, the shape parameter $a$ in the inverse-gamma distribution must be set to a reasonable value. For our model, the estimated group variance is always much larger than zero because the mean of estimated variance is approximately $\frac{a + J}{2}$ (Eq. \eqref{eq:cond_post_sigma_fin}) where $J$ is the number of groups. Therefore, $a$ can be set to a sufficiently large value, such as 2.







\subsection{Inference}

Given an observed count dataset structured using multiple groups, $\cD$, fitting an HBPRM entails the estimation of the joint posterior density distribution of all the parameters, which is only known up to a constant. If we denote the parameters by $ {\Theta} = \left\{     w _{11},\dots,     w _{jk},\dots,      w _{JK};\mu _{1},\dots,\mu _{{K}};\sigma_{1}^2,\dots,\sigma _{K}^2 \right\}$, then the joint posterior factorizes as
\begin{align}
    p\left( {\Theta}|y,     x \right)  \propto &\prod\limits_{j = 1}^J \prod\limits_{i = 1}^{{n_j}} {\rm{Pois}}\left( {{     y_{ij}} \mid \exp \left(\sum\limits_{k = 1}^K {{ w_{jk}}{ x_{ijk}}} \right) } \right) \times \nonumber\\
    & \prod\limits_{k = 1}^K {\cN\left( {{     w _{jk}}|{\mu _{{k}}},\sigma _{{k}}^2} \right)} \cN \left( {{\mu _{{k}}}|m,{\tau ^2}} \right)  \mathcal{IG}\left( {\sigma _{{k}}^2|\frac{a}{2}, \frac{b}{2}} \right) .
\end{align}

\noindent Sampling from the joint posterior becomes a challenging task as it does not admit a closed-form expression. While MCMC algorithms (e.g., the MH) can be used, the need to judiciously tune the step size for the desired acceptance rate often repel users from using this algorithm~\cite{graves2011automatic,holden2019mixing}. In comparison, the Gibbs sampler is more efficient and does not require any tuning of the proposal distribution, therefore it has been used for Bayesian inference in a wide range of applications~\cite{kass1998markov,pang2001estimation}. Classical Gibbs sampling requires that one can directly sample from the conditional posterior distribution of each parameter (or block of parameters), such as conditional conjugate posterior distributions. The full conditional posteriors for implementing the Gibbs sampler are
\begin{align}
    p\left( {{     w_{jk}}| - } \right) &\propto {\prod\limits_{i = 1}^{{n_j}} {{\rm{Pois}}\left( {{     y_{ij}} \mid \exp \left(\sum\limits_{k = 1}^K {{ w_{jk}}{ x_{ijk}}} \right) } \right)\cN\left( {{     w_{jk}}|{\mu _{{ k}}},\sigma _{{ k}}^2} \right)} } \label{eq:cond_post_coeff},\\
    p\left( {{\mu _{{ k}}}| - } \right) &\propto {\cN\left( {{     w_{1k},\dots,      w_{Jk}}|{\mu _{{ k}}},\sigma _{{ k}}^2} \right) \cN \left( {{\mu _{{ k}}}|m,{\tau ^2}} \right)} \label{eq:cond_post_mu_orig},\\
    p\left( {\sigma _{{ k}}^2| - } \right) &\propto {\cN\left( {{     w_{1k},\dots,      w_{Jk}}|{\mu _{{ k}}},\sigma _{{ k}}^2} \right)\mathcal{IG}\left( {\sigma _{{ k}}^2|\frac{a}{2},\frac{b}{2}} \right)}, \label{eq:cond_post_sigma_orig}
\end{align}

\noindent where $p\left(\cdot |-\right)$ represents the conditional posterior of a parameter of interest given the remaining parameters and the data. Due to the Gaussian-Gaussian and Gaussian-inverse-gamma conjugacy, Eq. \eqref{eq:cond_post_mu_orig} and Eq. \eqref{eq:cond_post_sigma_orig} can be expressed in an analytical form~\cite{murphy2007conjugate}
\begin{align}
    p\left( {{\mu _{{ k}}}| - } \right) &\propto \cN {\left( {{\mu _{{ k}}}\left| {\frac{1}{{\frac{1}{{{\tau ^2}}} + \frac{J}{{\sigma _{{ k}}^2}}}}\left( {\frac{m}{{{\tau ^2}}} + \frac{{\sum\limits_{j = 1}^J {{     w_{jk}}} }}{{\sigma _{{ k}}^2}}} \right)} \right.,\frac{1}{{\frac{1}{{{\tau ^2}}} + \frac{J}{{\sigma _{{ k}}^2}}}}} \right)}, \label{eq:cond_post_mu_fin}\\
    p\left( {\sigma _{{ k}}^2 | - } \right) &\propto \mathcal{IG} \left( \sigma _{{ k}}^2\left|\frac{a+J}{2}\right., \frac{b+\sum\limits_{j = 1}^J {\left(     w_{jk}-\mu_{k}\right)^2}}{2} \right). \label{eq:cond_post_sigma_fin} 
\end{align}

\noindent However, Eq.~\eqref{eq:cond_post_coeff} does not admit an analytical solution because the Poisson likelihood is not conjugate to the Gaussian prior. Consequently, it is challenging to sample directly from the conditional posterior to enable the Gibbs sampler. In this case, other algorithms can be used to obtain $p\left( {{     w_{jk}}| - } \right)$, such as adaptive-rejection sampling~\cite{gilks1992adaptive}, and Metropolis-within-Gibbs algorithm~\cite{geweke2001bayesian}. However, these algorithms introduce an additional computational cost due to the need to evaluate the complex conditional distribution. Therefore, we propose to use a Gaussian approximation to the Poisson likelihood given in Eq.~\eqref{eq:cond_post_coeff} to obtain a closed-form solution to the conditional posterior of coefficients. With the closed-form solution, the complex inference of regression coefficients can be simplified to save computational resources. Reducing the computational cost of sampling from $p\left( {{     w_{jk}}| - } \right)$ is critical for datasets with a large number of groups because the number of regression coefficients, $J \times K$, can be significantly larger than the number of prior parameters, $2K$.

\subsection{Gaussian Approximation to Log-gamma Distribution} \label{sec:gaus_approx}
This section introduces the Gaussian approximation to the log-gamma distribution that is used to obtain the closed-form approximate conditional posterior distribution in Section \ref{sec:approx_post}. Consider a gamma random variable $z$ with probability density function (pdf) given by
\begin{equation} \label{eq:pdf_gamma}
    p\left(z|\alpha,\beta\right) = \frac{z^{\alpha-1} e{-\frac{z}{\beta}}}{\Gamma\left(\alpha\right)\beta^{\alpha}},\; \alpha>0,\; \beta >0,
\end{equation}

\noindent where $\Gamma\left(\cdot\right)$ is the gamma function, and $\alpha$ and $\beta$ are the location parameter and the scale parameter, respectively.
The random variable, $\ln z \in \bR$, follows a log-gamma distribution. The mean $\mu_z$ and variance $\sigma^2_z$ of log-gamma distribution are calculated using Eq. \eqref{eq:log_gamma_mean} and Eq. \eqref{eq:log_gamma_var}, respectively~\cite{halliwell2018log}.
\begin{subequations}
\label{eq:log_gamma_mean}
\begin{align}
    \mu_z  &= {\psi_0}\left( \alpha \right) + \ln \beta  \label{eq:log_gamma_mean_a}\\
    &=  - \gamma  + \sum\limits_{n = 1}^\infty  {\left( {\frac{1}{n} - \frac{1}{{n + \alpha - 1}}} \right)} + \ln \beta \label{eq:log_gamma_mean_b}\\
    & =  - \gamma  + \sum\limits_{n = 1}^{\alpha - 1} {\frac{1}{n}} + \ln \beta \label{eq:log_gamma_mean_c}
\end{align}
\end{subequations}

\begin{subequations}
\label{eq:log_gamma_var}
\begin{align}
    \sigma^2_z & = {\psi_1}\left( \alpha \right) \label{eq:log_gamma_var_a}\\
    & = \sum\limits_{n = 0}^\infty  {\frac{1}{{{{\left( {\alpha + n} \right)}^2}}}} \\
    & = \frac{{{\pi ^2}}}{6} - \sum\limits_{n = 1}^{\alpha - 1} {\frac{1}{{{n^2}}}}
\end{align}
\end{subequations}


\noindent In Eq. \eqref{eq:log_gamma_mean_a} and Eq. \eqref{eq:log_gamma_var_a}, $\psi_0(\cdot)$ and $\psi_1(\cdot)$ are the zeroth and the first order of polygamma functions~\cite{batir2007some}. In Eq. \eqref{eq:log_gamma_mean_b} and Eq. \eqref{eq:log_gamma_mean_c}, $\gamma$ is the Euler-Mascheroni constant~\cite{mavcys2013euler}. 

For large values of $\alpha$, the pdf of log-gamma distribution can be approximated by that of a Gaussian distribution~\cite{chan2009bayesian, prentice1974log}, shown in Eq. \eqref{eq:Gaus_approx_log_gamma_alpha}.
\begin{equation} \label{eq:Gaus_approx_log_gamma_alpha}
    \textrm{Log-gamma}(\ln z | \alpha, \beta) \approx \cN\left(\ln z|{\psi_0}\left( \alpha \right) + \ln \beta,\, {\psi_1}\left( \alpha \right)\right)
\end{equation}

\noindent 
To apply the approximation in the conditional posterior (Eq. \eqref{eq:orig_likeli_wjk} to Eq. \eqref{eq:approx_likeli}), we need to include $y$ in the pdf of log-gamma distribution. Therefore, we let $\alpha=y$ and $\beta=1$, and Eq. \eqref{eq:Gaus_approx_log_gamma_alpha} becomes
\begin{equation} \label{eq:Gaus_approx_log_gamma_y}
    \textrm{Log-gamma}(\ln z | y,1) \approx \cN\left(\ln z|{\psi_0}\left( y \right),\, {\psi_1}\left( y \right)\right).
\end{equation}

\noindent
Note that because $\alpha > 0$ and $\alpha$ is replaced by count data $y$, the approximation can only be applied to positive counts.

Similarly, plugging in $\alpha=y$, $\beta=1$, and $\Gamma (n)=(n-1)!$ where $n \in \{1,2,3,\dots\}$, Eq. \eqref{eq:pdf_gamma} becomes

\vspace{-12pt}

\begin{equation} \label{eq:gamma_y_1}
    p\left(z|y,1\right) = \frac{z^{y-1} e^{-z}}{\left(y-1\right)!}.
\end{equation}

Next, we need to relate Eq. \eqref{eq:Gaus_approx_log_gamma_y} and Eq. \eqref{eq:gamma_y_1}. First, using the ``change of variable'' method and substituting $\ln z$ with $v$~\cite{chan2009bayesian} in Eq. \eqref{eq:gamma_y_1},  we obtain the pdf of $v$
\begin{subequations}
\begin{align}
    p\left(v|y,1\right) &= p\left(z=e^{v}|y,1\right) \frac{\partial e^v}{{\partial v}}\\
    &= \frac{1}{(y-1)!}e^{v y}e^{-e^{v}}. \label{eq:pdf_nu}
\end{align}
\end{subequations}

\noindent Then, using Eq. \eqref{eq:Gaus_approx_log_gamma_y} yields
\begin{equation} \label{eq:approx_pdf_nu}
    \frac{1}{(y-1)!}e^{v y}e^{-e^{v}} \approx \frac{1}{\sqrt{2 \pi\psi_1(y)} } e^ { \frac{(v-\psi_0(y))^2}{-2 \psi_1(y)}}.
\end{equation}

\vspace{-8pt}
\begin{figure}[htbp]
    \centering
    \includegraphics[width=1.01\textwidth]{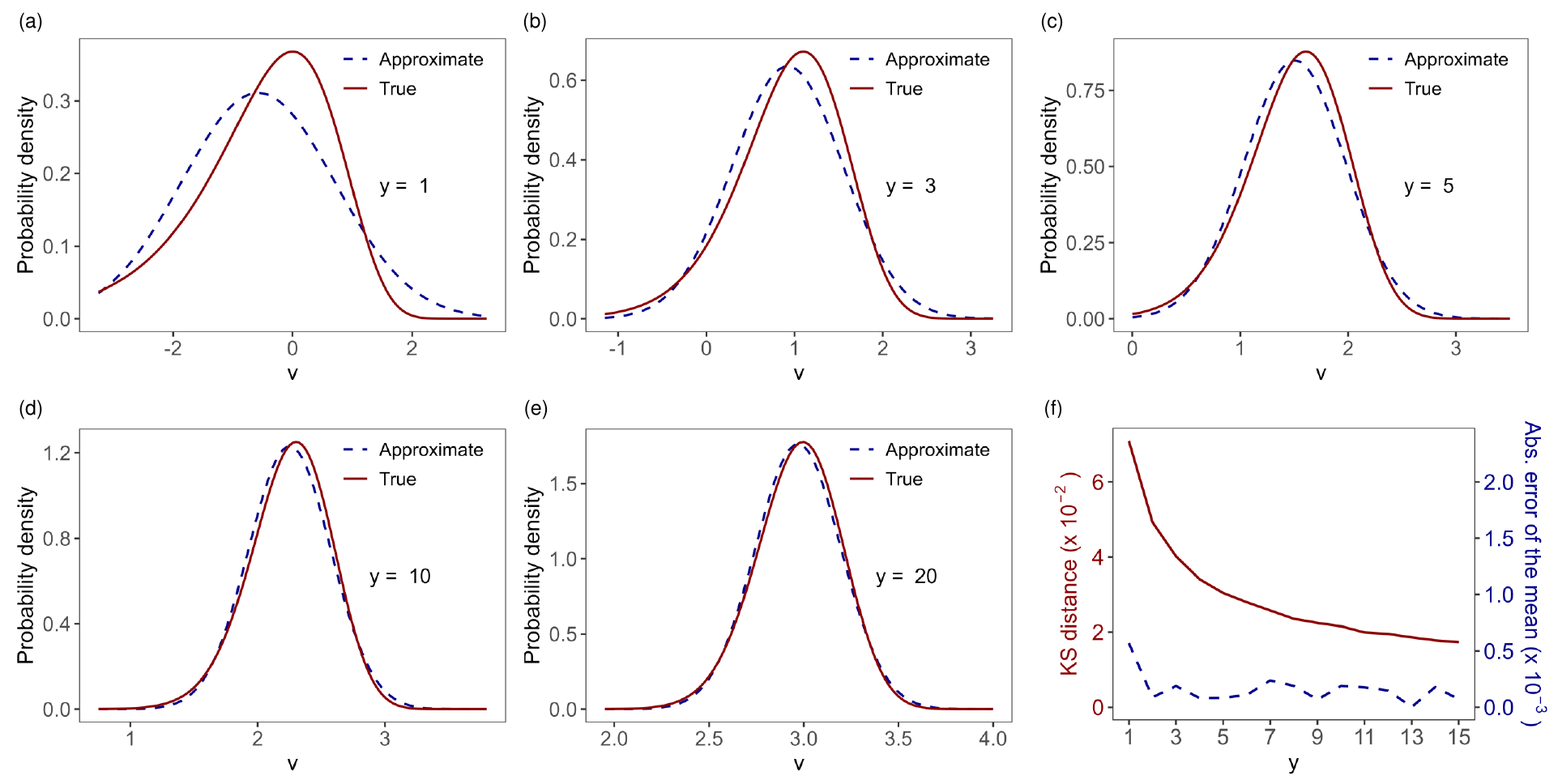}
    \vspace{-8pt}
    \caption{Quality of the Gaussian approximation (dashed blue line) to the true distribution (solid red line) for different values of $y$: (a) $y$=1; (b) $y$=3; (c) $y$=5; (d) $y$=10; (e) $y$=20. (f) Values of KS distance (solid red line) between the approximate distribution and true distribution and values of the absolute error in the mean (dashed blue line) of the approximate distribution as the value of $y$ increases.}
    \label{fig:approx_quality}
\end{figure}

The comparison between the true and approximate Gaussian distribution, i.e., left and right-hand sides of Eq. \eqref{eq:approx_pdf_nu} respectively, is shown in Fig. \ref{fig:approx_quality}. When the counts are small, such as $y \le 5$ ((a) to (c)), the approximation is not very close to the true distribution. We can also see that the Kolmogorov–Smirnov (KS) distance shown in Fig. \ref{fig:approx_quality} (f), defined as the largest absolute difference between the cumulative density functions for the approximate distribution and true distribution, is relatively large. As the value of $y$ increases, the approximate Gaussian distribution is increasingly closer to the true distribution. Also, the absolute error in the mean value of the Gaussian approximation is relatively small when $y$ is greater than 3. Notice again that the approximation given by Eq. \eqref{eq:approx_pdf_nu} is not directly applicable to zero counts. However, when zero counts are present in the dataset, such as in epidemiology studies, one can increase each count by a positive count (e.g., 5). This linear transformation allows one to circumvent the problem arising from zero counts (dependent variable) without compromising the model accuracy, which is because such a transformation does not change the distribution of the error and preserves the relation between the dependent and independent variables.

\subsection{Closed-form Approximate Conditional Posterior Distribution} \label{sec:approx_post}

In the conditional posterior of coefficient $ w_{jk}$ given by Eq. \eqref{eq:cond_post_coeff}, the likelihood function is
\begin{equation} \label{eq:orig_likeli_wjk} 
    {\prod\limits_{i = 1}^{{n_j}} {{\rm{Pois}}\left( {{     y_{ij}} \mid e^{\sum_{k = 1}^K {{ w_{jk}}{ x_{ijk}}} } } \right)}}  = {\prod\limits_{i = 1}^{{n_j}} {{\frac{1}{     y_{ij}(     y_{ij}-1)!} e^{\sum_{k=1}^{K} {w_{jk}      x_{ijk}}     y_{ij}}e^{- e^{\sum_{k=1}^{K} {w_{jk}      x_{ijk}}}}}}}.
\end{equation}

\noindent Applying the approximation given by Eq. \eqref{eq:approx_pdf_nu} yields
\begin{equation}\label{eq:approx_likeli}
    {\prod\limits_{i = 1}^{{n_j}} {{\rm{Pois}}\left( {{     y_{ij}} \mid e^{\sum_{k = 1}^K {{ w_{jk}}{ x_{ijk}}} } } \right)}}  \approx {\prod\limits_{i = 1}^{n_j} {{\frac{1}{     y_{ij}}} \frac{1}{ \sqrt{2 \pi\psi_1(y_{ij})} } e^ { \frac{\left({\sum_{k=1}^{K} {w_{jk}      x_{ijk}}}-\psi_0(y_{ij})\right)^2}{-2 \psi_1(y_{ij})}}}}. 
\end{equation}

\noindent Plugging Eq. \eqref{eq:approx_likeli} into Eq. \eqref{eq:cond_post_coeff} we get
\begin{align}
    p\left( {{     w _{jk}}| - } \right) &\propto \exp \left[ {\frac{{{{\left( {{     w _{jk}} - {\mu _{{k}}}} \right)}^2}}}{{ - 2\sigma _{{k}}^2}}} \right]\prod\limits_{i = 1}^{{n_j}} {\exp \left\{ {\frac{{{{\left[ {\sum\limits_{k = 1}^K {{ w_{jk}}{ x_{ijk}}}  - {\psi _0}\left( {{     y_{ij}}} \right)} \right]}^2}}}{{ - 2{\psi _1}\left( {{     y_{ij}}} \right)}}} \right\}}  \\
    & = \exp\left\{ {\frac{{{{\left( {{     w _{jk}} - {\mu _{{k}}}} \right)}^2}}}{{ - 2\sigma _{{k}}^2}} + \sum\limits_{i = 1}^{{n_j}} {\frac{{{{\left[ {\sum\limits_{k = 1}^K {{ w_{jk}}{ x_{ijk}}}  - {\psi _0}\left( {{     y_{ij}}} \right)} \right]}^2}}}{{ - 2{\psi _1}\left( {{     y_{ij}}} \right)}}} } \right\}.
\end{align}

\noindent As the product of two Gaussians is still Gaussian, the posterior can also be written as
\begin{equation}
    p\left( {     w _{jk}| - } \right) \propto \exp \left[ {\frac{{{{\left( {     w _{jk} - {{\widehat \mu }_{{{k}}}}} \right)}^2}}}{{-2 \widehat \sigma _{{{k}}}^2}}} \right] \label{eq:cond_post_coeff_fin},
\end{equation}

\noindent where ${\widehat \mu }_{{{k}}}$ and ${\widehat \sigma _{{{k}}}^2}$ are the mean and variance of the approximate Gaussian posterior. Completing the squares (see Appendix \ref{app:derivation} for more details) we get
\begin{align}
    {\widehat \mu }_{{{k}}} &= \frac{{{\mu _{{k}}} + \sigma _{{k}}^2\sum\limits_{i = 1}^{{n_j}} {\frac{{{     x_{ijk}}}}{{{\psi _1}\left( {{     y_{ij}}} \right)}}} \left[ {{\psi _0}\left( {{     y_{ij}}} \right) - \sum\limits_{h = 1,h \ne k}^K {{     x_{ijh}}{w _{jh}}} } \right]}}{{\sigma _{{k}}^2\sum\limits_{i = 1}^{{n_j}} {\frac{{     x_{ijk}^2}}{{{\psi _1}\left( {{     y_{ij}}} \right)}}}  + 1}},\, \forall j=1,\dots,J, \\
    {\widehat \sigma _{{{{k}}}}^2} &= \frac{{\sigma _{{k}}^2}}{{\sigma _{{k}}^2\sum\limits_{i = 1}^{{n_j}} {\frac{{     x_{ijk}^2}}{{{\psi _1}\left( {{     y_{ij}}} \right)}}}  + 1}}, \, \forall j=1,\dots,J.
\end{align}

Now that the full conditional posterior distributions can be expressed analytically, we can construct the approximate Gibbs sampler (Algorithm~\ref{alg:ags}) to obtain posterior samples of the parameters in HBPRM efficiently.
\begin{algorithm}
\caption{Approximate Gibbs sampler}\label{alg:ags}
\begin{algorithmic}[1]
\Input{$ {x}$, $ {y}$, number of samples as warm-ups $N_0$, number of desired samples $N_1$}.
\Output{Desired posterior samples, $\mu_{k}^{(\ell)},\sigma_{k}^{2(\ell)},     w_{jk}^{(\ell)}$,  $\ell=N_0+1,\dots,N_0+N_1$, $k=1,\dots,K$, and $j=1,\dots,J$}.
\State Generate the initial sample $\mu_{k}^{(0)},\sigma_{k}^{2(0)},     w_{jk}^{(0)}$.
\For{$\ell=1$ to $(N_0+N_1)$}{}
    \For{$k=1$ to $K$}{}
        \State Sample hyperparameter $\mu_{k}^{(\ell)}$ according to Eq. \eqref{eq:cond_post_mu_fin}.
        \State Sample hyperparameter $\sigma_{k}^{2(\ell)}$ according to Eq. \eqref{eq:cond_post_sigma_fin}.
        \For{$j=1$ to $J$}{}
            \State Sample each parameter $     w_{jk}^{(\ell)}$ according to Eq. \eqref{eq:cond_post_coeff_fin}.
        \EndFor
    \EndFor
\EndFor
\end{algorithmic}
\end{algorithm}



\section{Experiments} \label{sec:experiments}

We evaluate the performance of our proposed AGS algorithm by applying it to several synthetic and real data sets. The performance of AGS is evaluated in terms of the accuracy, efficiency, and computational time. The proposed approach is compared with the state-of-the-art MCMC algorithm, No-U-Turn sampler (NUTS)~\cite{hoffman2014no}, using the same datasets and performance metrics. NUTS is an extension to Hamiltonian Monte Carlo (HMC) algorithm that exploits Hamiltonian dynamics to propose samples. NUTS can free users from tuning the proposals and has been demonstrated to provide efficient inference of complex hierarchical models~\cite{betancourt2015hamiltonian}. The description of the datasets and the experimental setup is provided in this section. The code and non-confidential data used for the experiments are available on the GitHub account of the corresponding author. 

\subsection{Data Description}

Multiple synthetic and real datasets are used to evaluate the performance of AGS for different data types and sizes. This section describes the approach to generating synthetic data and the characteristics of real datasets which include power outages, Covid-19 positive cases, and bike rentals. A subset of each dataset is provided in tables \ref{tab:syn_data} to \ref{tab:bike_share}.

\textit{Synthetic data}. The synthetic datasets are generated according to the model shown in Eq. \eqref{eq:syn_data}. This model ensures that the generated datasets contain a specified range of counts and closely mimic the number of emergency incidents during disasters, such as the number of power outages after a severe storm. An example of the synthetic dataset is presented in Table \ref{tab:syn_data}. Note that the data for all $x_k$, $k=1,...,6$, in each group is sorted in ascending order before calculating $y$ to ensure a more consistent relationship between $y$ and $x$.
\begin{subequations} \label{eq:syn_data}
    \begin{align}
    {x_{ij1}} &\sim U\left( {0.1,\,2} \right), \;i=1,\dots, n_j,\;j=1,\dots,J\\
    {x_{ij2}} &\sim U\left( {0.1,\,1} \right), \;i=1,\dots, n_j,\;j=1,\dots,J\\
    {x_{ij3}} &\sim U\left( {0.1,\,0.5} \right), \;i=1,\dots, n_j,\;j=1,\dots,J\\
    {x_{ij4}} &\sim U\left( {1,\,10} \right), \;i=1,\dots, n_j,\;j=1,\dots,J\\
    {x_{ij5}} &\sim U\left( {0.5,\,5} \right), \;i=1,\dots, n_j,\;j=1,\dots,J\\
    {x_{ij6}} &\sim U\left( {10,\,100} \right), \;i=1,\dots, n_j,\;j=1,\dots,J\\
    x_{\cdot j} &\sim U\left( {{{10}^4},\,{{10}^6}} \right),\;j=1,\dots,J\\
    {     w _{jk}} &\sim \mathcal{N}\left( {0.001, \,0.001} \right),\; j = 1,\dots,J,\; k=1,\dots, K\\
         y_{ij} &= {\left( e^{ \sum_{k=1}^K      w _{jk}      x_{ijk}} \right)_{{\rm{min-max}}}}x_{\cdot j},\;i=1,\dots, n_j,\;j=1,\dots,J.
    \end{align}
\end{subequations}

In Eq. \eqref{eq:syn_data}, 
the notation $\left(\cdot\right)_{{\rm{min-max}}}$ represents the min-max normalizing function\footnote{For an array of real numbers represented by a generic vector $ {\bx}$. The min-max normalization of $ {\bx}$ is given by $ {\bx}_{\rm{min-max}} = \frac{ {\bx}-\bx_{\rm{min}}}{\bx_{\rm{max}}-\bx_{\rm{min}}}$.}. Each count $  y_{ij}$ is rounded to the nearest integer. $x_{\cdot j}$ is the group-level covariate for group $j$. We generate 15 synthetic datasets (S1, \dots, S15) with varying total numbers of data points ($N_d$) in each dataset and varying numbers of data points in each group $n_j$ (for simplicity, it is assumed the same for each group in the same synthetic dataset), numbers of covariates $K$ ($K \le 6$), and numbers of groups $J$ to analyze the effect of the size of the data on the performance of AGS and NUTS (Table~\ref{tab:results_simulated_datasets}).

\begin{table}[!htb]
  \centering
  \caption{An example of synthetic dataset}
    \begin{tabular*}{\textwidth}{c @{\extracolsep{\fill}}cccccc}
    \toprule
    \multicolumn{1}{l}{~$x_1$} & \multicolumn{1}{l}{$x_2$} & \multicolumn{1}{l}{~$x_3$} & \multicolumn{1}{l}{~$x_4$} & \multicolumn{1}{l}{~~$x_5$} & \multicolumn{1}{l}{~~$x_6$} & \multicolumn{1}{l}{~~~$y$} \\
    \hline
    0.14  & 0.11  & 0.27  & 1.49  & 0.66  & 12.10 & 42 \\
    0.61  & 0.15  & 0.27  & 2.22  & 1.52  & 24.76 & 6440 \\
    0.64  & 0.58  & 0.27  & 3.11  & 1.93  & 34.67 & 11424 \\
    0.77  & 0.58  & 0.30  & 5.70  & 2.13  & 38.71 & 13535 \\
    1.07  & 0.60  & 0.34  & 6.38  & 2.77  & 62.73 & 25064 \\
    1.29  & 0.75  & 0.42  & 7.78  & 3.69  & 79.38 & 33917 \\
    1.41  & 0.91  & 0.47  & 8.84  & 4.91  & 85.56 & 38272 \\
    1.55  & 0.93  & 0.49  & 8.93  & 4.96  & 92.86 & 41806 \\

    \bottomrule
    \end{tabular*}%
  \label{tab:syn_data}%
\end{table}%

\textit{Power outage data}. The power outage data includes the number of customers without power in multiple counties following 11 disruptive events (denoted by P1$,\dots,$ P11). The power outage dataset following a particular disruptive event is grouped by county, i.e., power outage counts for the same county after a particular disruptive event fall into the same group. The covariates in each dataset include PS (surface pressure, Pa), TQV (precipitable water vapor, kg$\cdot \textrm{m}^{-2}$), U10M (10-meter eastward wind speed, m/s), V10M (10-meter northward wind speed, m/s), $t$ (time after the start of an event, hours). The outage datasets were collected from public utility companies during severe weather events and the weather data from the National Oceanic and Atmospheric Administration.

\begin{table}[!htb]
  \centering
  \caption{A subset of power outage data}
    \begin{tabular*}{\textwidth}{c @{\extracolsep{\fill}}cccccc}
    \toprule
    \multicolumn{1}{l}{Event ID} & \multicolumn{1}{l}{~~PS} & \multicolumn{1}{l}{TQV} & \multicolumn{1}{l}{U10M} & \multicolumn{1}{l}{V10M} & \multicolumn{1}{l}{$\;\;t$} & \multicolumn{1}{l}{Outage count}\\
    \hline
    1     & 99691.45 & 43.18 & 2.39  & 4.79  & 4     & 66807 \\
    1     & 100917.62 & 26.75 & 1.11  & 1.39  & 32    & 18379 \\
    1     & 101041.88 & 36.29 & 1.11  & -0.18 & 60    & 12096 \\
    1     & 101467.45 & 55.72 & -1.73 & -0.67 & 116   & 14231 \\
    1     & 101155.43 & 37.50 & -0.08 & -3.01 & 144   & 10155 \\
    1     & 101037.79 & 32.13 & -0.48 & -1.53 & 172   & 4758 \\
    1     & 101194.86 & 40.20 & -0.90 & 1.66  & 200   & 2699 \\
    1     & 101183.98 & 46.61 & -0.54 & 1.20  & 228   & 2297 \\
    1     & 101136.76 & 34.68 & -1.14 & -1.50 & 256   & 248 \\
    1     & 101086.31 & 43.80 & -2.19 & -2.44 & 284   & 43 \\
    \bottomrule
    \end{tabular*}%
    \label{tab:power_outage}
\end{table}%

\textit{Covid-19 test data}. The Covid-19 test dataset is obtained from Ref. \cite{kucirka2020variation}, which are originally collected from seven papers (two preprints and five peer-reviewed articles) that provide data on RT-PCR (reverse transcriptase polymerase chain reaction) performance by time since the symptom onset or exposure using samples derived from nasal or throat swabs among patients tested for Covid-19. The number of studies (groups) is 11. Each study includes multiple test cases (Table \ref{tab:covid}), each of which includes the days, $t$, after exposure to Covid-19, and the total number of samples tested, $N_s$. The response variable is the number of patients who tested positive among the samples. The total number of test cases is 379. As the proposed approximation cannot be applied to zero counts, we remove the test cases with zero positive test among the samples tested. The total number of test cases after removing those with zero counts is 298. The test cases are grouped by studies. Following Ref. \cite{kucirka2020variation}, the exposure is assumed to have occurred five days before the symptom onset and $\log (t)$, $\log (t)^2$, $\log (t)^3$, and $N_s$ are used as the covariates.
\begin{table}[!htb]
  \centering
  \caption{A subset of the Covid-19 test data}
    \begin{tabular*}{\textwidth}{c @{\extracolsep{\fill}}cccccc}
        \toprule
    \multicolumn{1}{l}{Study ID}
    & \multicolumn{1}{l}{Test case ID}
    & \multicolumn{1}{l}{$\log(t)$} & \multicolumn{1}{l}{$\log (t)^2$} & \multicolumn{1}{l}{$\log (t)^3$} &
    \multicolumn{1}{l}{$N_s$}&
    \multicolumn{1}{l}{Positive count} \\
    \hline
    1     & 1     & 1.23  & 1.51  & 1.86  & 35    & 15 \\
    1     & 2     & 1.26  & 1.58  & 1.98  & 23    & 11 \\
    1     & 3     & 1.28  & 1.64  & 2.09  & 20    & 6 \\
    1     & 5     & 1.32  & 1.75  & 2.31  & 20    & 8 \\
    1     & 6     & 1.34  & 1.80   & 2.42  & 11    & 3 \\
    1     & 7     & 1.36  & 1.85  & 2.53  & 11    & 5 \\
    1     & 8     & 1.38  & 1.90   & 2.63  & 9     & 2 \\
    1     & 9     & 1.40   & 1.95  & 2.73  & 6     & 3 \\
    1     & 10    & 1.41  & 1.98     & 2.83  & 5     & 2 \\
    \bottomrule
    \end{tabular*}%
    \label{tab:covid}
\end{table}%

\textit{Bike sharing data}. The bike sharing data include daily bike rental counts for 729 days and the covariates we use include normalized temperature, normalized humidity, and casual bike rentals. The dataset is obtained from the UCI Machine Learning Repository~\cite{asuncion2007uci}. Bike rental counts are grouped by whether the rental occurs on a working day (Table \ref{tab:bike_share}).

\begin{table}[!htb]
  \centering
  \caption{A subset of the bike sharing data. Workingday=1 indicates the rental occurs on a working day, and 0 otherwise.}
    \begin{tabular*}{\textwidth}{c @{\extracolsep{\fill}} cccc}
    \toprule
    \multicolumn{1}{l}{Workingday} & \multicolumn{1}{l}{Temperature} & \multicolumn{1}{l}{Humidity} & \multicolumn{1}{l}{Casual rental count} & \multicolumn{1}{l}{Total count} \\
    \hline
    0     & 0.34  & 0.81  & 331   & 985 \\
    0     & 0.36  & 0.70  & 131   & 801 \\
    1     & 0.20  & 0.44  & 120   & 1349 \\
    1     & 0.20  & 0.59  & 108   & 1562 \\
    1     & 0.23  & 0.44  & 82    & 1600 \\
    1     & 0.20  & 0.52  & 88    & 1606 \\
    1     & 0.20  & 0.50  & 148   & 1510 \\
    0     & 0.17  & 0.54  & 68    & 959 \\
    0     & 0.14  & 0.43  & 54    & 822 \\
    \bottomrule
    \end{tabular*}%
  \label{tab:bike_share}%
\end{table}%

To investigate the performance of AGS for small counts, including zero counts, we also simulate datasets with small counts using the following model shown in Eq. \eqref{eq:syn_data_small}.

\begin{subequations} \label{eq:syn_data_small}
    \begin{align}
    {x_{ij1}} &\sim U\left( {0.1,\,2} \right), \;i=1,\dots, n_j,\;j=1,\dots,J\\
    {x_{ij2}} &\sim U\left( {0.1,\,1} \right), \;i=1,\dots, n_j,\;j=1,\dots,J\\
    {x_{ij3}} &\sim U\left( {0.1,\,0.5} \right), \;i=1,\dots, n_j,\;j=1,\dots,J\\
    {x_{ij4}} &\sim U\left( {1,\,10} \right), \;i=1,\dots, n_j,\;j=1,\dots,J\\
    {x_{ij5}} &\sim U\left( {0.5,\,5} \right), \;i=1,\dots, n_j,\;j=1,\dots,J\\
    x_{\cdot j} &\sim TEXP\left(0.7, \, 1, \, y_{\max}\right),\;j=1,\dots,J\\
    {     w _{jk}} &\sim \mathcal{N}\left( {0.1, \,0.1} \right),\; j = 1,\dots,J,\; k=1,\dots, K\\
         y_{ij} &= \left\lfloor {\left( e^{ \sum_{k=1}^K      w _{jk}      x_{ijk}} \right)_{{\rm{min-max}}}}x_{\cdot j} \right\rfloor,\;i=1,\dots, n_j,\;j=1,\dots,J. \label{eq:y_small}
    \end{align}
\end{subequations}

\noindent
$TEXP$ represents the truncated exponential distribution. The PDF of $TEXP(0.7,\, 1,\, y_{\max})$, where 0.7 is the rate parameter while 1 and $y_{\max}$ are the lower and upper bounds, is given by

\begin{equation}
    f(x) = \begin{cases}
0.7  e^{ - 0.7 ( y_{\max} - x - 1)}, & x \in [1,\, y_{\max}], \\
0, & \textrm{otherwise}.
\end{cases}
\end{equation}

\noindent This particular truncated exponential distribution instead of a uniform distribution is used to ensure that the generated counts will not concentrate on small values. By changing the value of the upper bound, we can generate counts in different ranges. Notice that since the floor function, $\left\lfloor \cdot \right\rfloor$, is used in Eq. \eqref{eq:y_small}, the generated counts can have zeros, which is smaller than the lower bound 1.


\subsection{Experiment Setup}

In the HBPRM for the count datasets listed above, we employ $\cN(0,1)$ and $\mathcal{IG}\left(1, 1\right)$ as a weakly-informative prior~\cite{stan2016stan} for $\mu_k$ and $\sigma^2_k$ respectively. In the numerical experiments, NUTS is implemented with Stan~\cite{stan2016stan}. Numbers are averaged over 4 runs of 10000 iterations for each algorithm, discarding the first 5000 samples as warm-ups/burn-ins. We compare AGS with NUTS in terms of average sampling time in seconds per 1000 iterations ($T_s$), sampling efficiency ($E_s$), $R^2$, and Root Mean Square Error (RMSE). Sampling efficiency is quantified as the mean effective sampler size ($\hat{n}_{\mathrm{eff}}$) over the average sampling time in seconds per 1000 iterations, i.e., $E_s = \hat{n}_{\mathrm{eff}} / T_s$, where $\hat{n}_{\mathrm{eff}}$ is the effective sample size of multiple sequences of samples~\cite[Chapter 11]{gelman2013bayesian}. To make this paper self-contained, we have included the details for calculating $\hat{n}_{\mathrm{eff}}$, $R^2$, and RMSE in Appendix~\ref{app:metrics}. All experiments are implemented with \texttt{R} (version 3.6.1) on a Windows 10 desktop computer with a 3.40 GHz Intel Core i7-6700 CPU and 16.0 GB RAM.

\subsection{Results} \label{sec:results}
 
The performances of NUTS and AGS under different datasets are summarized in Table~\ref{tab:results_simulated_datasets} (synthetic datasets) and Table~\ref{tab:results_real_datasets} (real datasets). On both the synthetic and real datasets, AGS consistently outperforms NUTS in the average sampling time, especially when the size of the datasets is large. Depending on the dataset, the improvement in the average inference speed can be greater than one order of magnitude. This observation shows that using the Gaussian approximation to avoid the evaluation of complex conditional posterior can significantly boost the sampling speed. However, for all the datasets except for power outage dataset P1, the sampling efficiency of AGS is significantly lower than that of NUTS because the effective sample size obtained from AGS is much lower than that from NUTS. The relatively low inference efficiency of AGS does not compromise the accuracy of parameter estimates. In all the examined datasets, $R^2$ and RMSE have comparable values for both AGS and NUTS. The inference accuracy is better (higher $R^2$ and lower RMSE) for AGS across all the synthetic datasets except for S14 where the RMSE of NUTS is marginally smaller than that of AGS. In eight out of the thirteen (about 62\%) real datasets, AGS has slightly higher $R^2$ and lower RMSE. In particular, the results on the Covid test data show that as long as a significant percentage of counts are not all very small counts, then the proposed approximate Gibbs sampler can outperform the NUTS in predictive accuracy. Overall, it can be concluded that AGS significantly decreases the computational load by allowing for faster sampling without compromising the accuracy of the estimates.

\begin{table}[!h]
  \centering
  \footnotesize
  \caption{Performance of NUTS and AGS on synthetic datasets}
    \begin{tabular*}{\textwidth}{c @{\extracolsep{\fill}}ccccccccccc}
    \toprule
    \multirow{2}[2]{*}{Dataset\;} & \multicolumn{3}{c}{Characteristics} & \multicolumn{2}{c}{$T_s$ (s)} & \multicolumn{2}{c}{$E_s$} & \multicolumn{2}{c}{$R^2$} & \multicolumn{2}{c}{RMSE}\\ \cmidrule{2-12} 
       & $N_d$ & $K$ & $J$ & NUTS & \multicolumn{1}{p{1.5em}}{AGS} & NUTS & \multicolumn{1}{p{1.25em}}{AGS} & NUTS & \multicolumn{1}{p{2.em}}{AGS} & NUTS & \multicolumn{1}{p{2.em}}{AGS}\\    
    \hline
    S1 & 200  & 2  & 10  & 1.51  & \textbf{0.96 } & \textbf{649.28 } & 33.46  & 0.9390  & \textbf{0.9450 } & 6500  & \textbf{6191 } \\
    S2 & 400  & 2  & 10  & 2.66  & \textbf{0.98 } & \textbf{373.46 } & 34.99  & 0.9430  & \textbf{0.9500 } & 5823  & \textbf{5455 } \\
    S3 & 800  & 2  & 20  & 5.05  & \textbf{1.63 } & \textbf{195.21 } & 18.63  & 0.9576  & \textbf{0.9626 } & 3526  & \textbf{3310 } \\
    S4 & 200  & 3  & 10  & 5.76  & \textbf{1.25 } & \textbf{170.39 } & 5.28  & 0.9614  & \textbf{0.9660 } & 4235  & \textbf{3976 } \\
    S5 & 400  & 3  & 10  & 16.63  & \textbf{1.29 } & \textbf{59.78 } & 2.03  & 0.9683  & \textbf{0.9710 } & 3450  & \textbf{3305 } \\
    S6 & 800  & 3  & 20  & 40.53  & \textbf{2.44 } & \textbf{24.30 } & 1.80  & 0.9677  & \textbf{0.9719 } & 3993  & \textbf{3728 } \\
    S7 & 200  & 4  & 10  & 7.64  & \textbf{1.63 } & \textbf{129.25 } & 1.59  & 0.9716  & \textbf{0.9760 } & 2955  & \textbf{2718 } \\
    S8 & 400  & 4  & 10  & 23.72  & \textbf{1.94 } & \textbf{41.69 } & 1.50  & 0.9639  & \textbf{0.9701 } & 4406  & \textbf{4012 } \\
    S9 & 800  & 4  & 20  & 45.00  & \textbf{3.31 } & \textbf{21.99 } & 0.49  & 0.9740  & \textbf{0.9770 } & 2958  & \textbf{2790 } \\
    S10 & 200  & 5  & 10  & 12.12  & \textbf{2.06 } & \textbf{81.26 } & 0.70  & 0.9574  & \textbf{0.9631 } & 4941  & \textbf{4593 } \\
    S11 & 400  & 5  & 10  & 24.70  & \textbf{2.41 } & \textbf{39.68 } & 0.49  & 0.9782  & \textbf{0.9826 } & 2514  & \textbf{2245 } \\
    S12 & 800  & 5  & 20  & 64.00  & \textbf{4.12 } & \textbf{15.44 } & 0.27  & 0.9767  & \textbf{0.9804 } & 3446  & \textbf{3157 } \\
    S13 & 200  & 6  & 10  & 14.66  & \textbf{2.44 } & \textbf{67.49 } & 0.23  & 0.9817  & \textbf{0.9876 } & \textbf{2720}  & 2721 \\
    S14 & 400  & 6  & 10  & 42.01  & \textbf{2.63 } & \textbf{20.17 } & 0.24  & 0.9922  & \textbf{0.9948 } & 1339  & \textbf{1128 } \\
    S15 & 800  & 6  & 20  & 93.20  & \textbf{4.97 } & \textbf{8.44 } & 0.17  & 0.9910  & \textbf{0.9940 } & 1994  & \textbf{1629 } \\
    \bottomrule
    \end{tabular*}%
  \label{tab:results_simulated_datasets}%
\end{table}%

\begin{figure}[!htb]
    \centering
    \subfloat[]{{\includegraphics[width=0.5\textwidth]{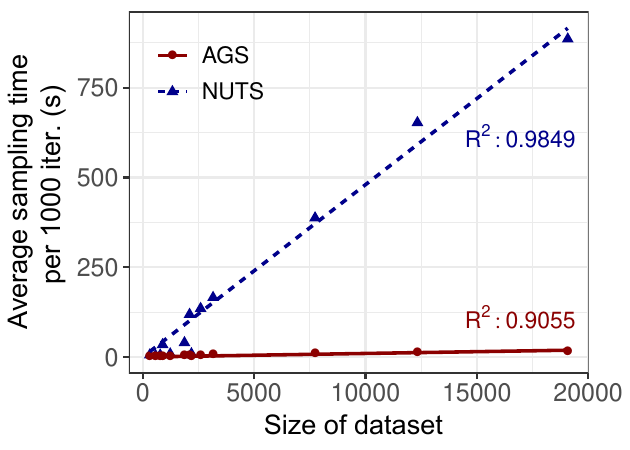} }}%
    \subfloat[]{{\includegraphics[width=0.5\textwidth]{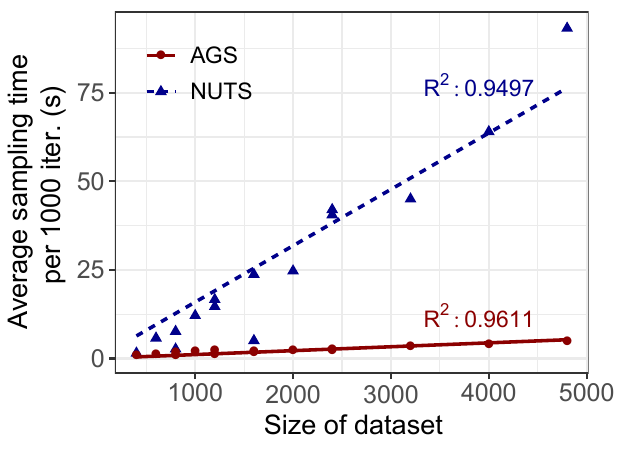} }}%
    \caption{Scalability of NUTS and AGS on (a) real datasets and (b) synthetic datasets. The size of a dataset, i.e., the total number of count data points, can be calculated by $\sum_{j=1}^{J} n_j$.}%
    \label{fig:scalability}%
\end{figure}

We also investigate the scalability of the two algorithms as it is crucial for large-scale hierarchical data. Therefore, we show the average sampling time of both algorithms across different dataset sizes to understand their performance for larger datasets. We conduct an empirical analysis of the average sampling time (seconds) per 1000 iterations for all the synthetic and real datasets shown in Fig.~\ref{fig:scalability}. The sampling time of both samplers increases as a function of the size of the dataset. However, when compared to NUTS, the increase in the sampling time of AGS is significantly lower, showing a significantly smaller rate of time increase over the size of datasets and suggesting improved scalability. This observation also indicates that although NUTS can generate samples effectively, it becomes inefficient in the case of large datasets as evaluating the gradient in proposing new samples becomes computationally expensive~\cite{nishio2019performance}.

\begin{table}[!h]  
  \centering
  \footnotesize
  \caption{Performance of NUTS and AGS on real datasets}
    \begin{tabular*}{\textwidth}{c @{\extracolsep{\fill}}ccccccccccc}
    \toprule
    \multirow{2}[2]{*}{Dataset\;} & \multicolumn{3}{c}{Characteristics} & \multicolumn{2}{c}{$T_s$ (s)} & \multicolumn{2}{c}{$E_s$} & \multicolumn{2}{c}{$R^2$} & \multicolumn{2}{c}{RMSE}\\ \cmidrule{2-12}
       & $N_d$ & $K$ & $J$ & NUTS & \multicolumn{1}{p{1.5em}}{AGS} & NUTS & \multicolumn{1}{p{1.25em}}{AGS} & NUTS & \multicolumn{1}{p{2.em}}{AGS} & NUTS & \multicolumn{1}{p{2.em}}{AGS}\\
    \hline
    P1 & 3817 & 5 & 56 & 885.65  & \textbf{15.76} & 1.12  & \textbf{1.25} & 0.9730  & \textbf{0.9801} & 13558 & \textbf{11621} \\
    P2 & 2467 & 5 & 50 & 652.87  & \textbf{13.71} & \textbf{1.50} & 0.62  & 0.9873  & \textbf{0.9884} & 1446  & \textbf{1384} \\
    P3 & 1548 & 5 & 35 & 387.47  & \textbf{9.41} & \textbf{2.54} & 0.92  & 0.9850  & \textbf{0.9870} & 1974  & \textbf{1833} \\
    P4 & 632 & 5 & 26 & 165.73  & \textbf{6.67} & \textbf{5.94} & 0.46  & \textbf{0.9923} & 0.9918  & \textbf{1327} & 1373  \\
    P5 & 520 & 5 & 16 & 135.16 & \textbf{4.31} & \textbf{7.27} & 1.85  & 0.9934  & \textbf{0.9940} & 3473  & \textbf{3312} \\
    P6 & 421 & 5 & 17 & 118.73 & \textbf{4.54} & \textbf{8.25} & 2.68  & 0.9908  & \textbf{0.9918} & 2526  & \textbf{2387} \\
    P7 & 375 & 5 & 23 & 39.86 & \textbf{5.76} & \textbf{24.75} & 2.41  & \textbf{0.9459} & 0.9355  & \textbf{3574} & 3903  \\
    P8 & 247 & 5 & 10 & 8.75  & \textbf{2.78} & \textbf{111.91} & 2.38  & \textbf{0.9744} & 0.9729  & \textbf{795} & 818\\
    P9 & 157 & 5 & 8  & 5.48  & \textbf{2.24} & \textbf{179.87} & 4.07  & 0.9964  & \textbf{0.9967} & 803  & \textbf{766} \\
    P10 & 115 & 5 & 6  & 4.49  & \textbf{1.72} & \textbf{218.17} & 6.75  & 0.9869  & \textbf{0.9915} & 7715  & \textbf{6222} \\
    P11 & 63 & 5 & 4  & 5.49  & \textbf{1.25} & \textbf{177.38} & 0.92  & \textbf{0.9356} & 0.9027  & \textbf{251} & 308  \\
    Bike share\; & 729 & 3 & 2  & 9.09  & \textbf{0.98} & \textbf{109.54} & 24.75  & \textbf{0.6743} & 0.6292  & \textbf{1101} & 1175 \\
    Covid test\; & 298 & 3 & 11  & 34.60  & \textbf{2.59} & \textbf{28.57} & 18.54  & 0.8517 &\textbf{0.8582}  & 2.53 &\textbf{2.47} \\
    \bottomrule
    \end{tabular*}%
  \label{tab:results_real_datasets}
\end{table}%


\begin{table}[!ht]
  \centering  
  \small
  \caption{Inference accuracy of NUTS and AGS on selected Covid-19 test datasets with small counts. The number of covariates $K=3$ and the number of groups $J=11$.}
    \begin{tabular}{lcccccccc}
    \toprule
    \multirow{2}[4]{*}{Dataset} & \multicolumn{4}{c}{Characteristics} & \multicolumn{2}{c}{$R^2$} & \multicolumn{2}{c}{RMSE} \\
\cmidrule{2-9}          & $RG$ & $PCT_0$ (\%)    & $PCT_{1,5}$ (\%)   & $N_d$  & NUTS  & AGS   & NUTS  & AGS \\
    \midrule
    Subset1 & [1, 5] & 0.0     & 100.0   & 168   & 0.4403 & \textbf{0.4452} & 0.9964 & \textbf{0.9920} \\
    Subset2 & [0, 5] & 32.5 & 67.5 & 249   & \textbf{0.6606} & 0.5684 & \textbf{0.9370} & 1.0570 \\
    Subset3 & [1, 38] & 0.0     & 56.4 & 298   & 0.8517 & \textbf{0.8582} & 2.5337 & \textbf{2.4685} \\
    Whole set & [0, 38] & 21.4 & 44.3 & 379   & \textbf{0.8692} & 0.8546 & \textbf{2.3492} & 2.4769 \\
    \bottomrule
    \end{tabular}%
  \label{tbl:covid_small}
\end{table}%

\begin{table}[htb]
  \centering \small
  \caption{Inference accuracy of NUTS and AGS on synthetic datasets with small counts. The number of covariates $K=5$ and the number of groups $J=8$.}
    \begin{tabular}{cccccrccc}
    \toprule
    \multirow{2}[3]{*}{Dataset} & \multicolumn{4}{c}{Characteristics} & \multicolumn{2}{c}{$R^2$} & \multicolumn{2}{c}{RMSE} \\
\cmidrule{2-9}          & \multicolumn{1}{c}{$RG$} & $PCT_0$ (\%) & $PCT_{1,5}$ (\%) & $N_d$  & NUTS & AGS   & NUTS  & AGS \\
\midrule
    SS1   & [1, 5] & 0.0  & 100.0  & 252   & \multicolumn{1}{c}{\textbf{0.9226}} & 0.8139 & \textbf{0.3198} & 0.4959 \\
    SS2   & [0, 5] & 21.3   & 78.8 & 320   & \multicolumn{1}{c}{\textbf{0.5414}} & 0.4567 & \textbf{0.9460} & 1.0010 \\
    SS3   & [1, 10] & 0.0   & 71.2  & 288   & \multicolumn{1}{c}{\textbf{0.9625}} & 0.9421 & \textbf{0.4478} & 0.5564 \\
    SS4   & [0, 10] & 10.0  & 64.1  & 320   & \multicolumn{1}{c}{0.6461} & \textbf{0.6480} & 1.4805 & \textbf{1.4764} \\
    SS5   & [1, 15] & 0.0   & 57.3  & 307   & \multicolumn{1}{c}{\textbf{0.9766}} & 0.9732 & \textbf{0.5715} & 0.6113 \\
    SS6   & [0, 15] & 4.1   & 55.0  & 320   & \multicolumn{1}{c}{\textbf{0.9641}} & 0.9538 & \textbf{0.7225} & 0.8193 \\
    SS7   & [1, 20] & 0.0   & 47.0  & 319   & \multicolumn{1}{c}{\textbf{0.9822}} & 0.9819 & \textbf{0.6639} & 0.6708 \\
    SS8   & [0, 20] & 3.1   & 46.9  & 320   & \multicolumn{1}{c}{\textbf{0.9691}} & 0.9642 & \textbf{0.8774} & 0.9446 \\
    SS9   & [1, 30] & 0.0   & 32.2  & 314   & \multicolumn{1}{c}{0.9723} & \textbf{0.9767} & 1.3059 & \textbf{1.1979} \\
    SS10  & [0, 30] & 1.9   & 31.6  & 320   & 0.9525 & \textbf{0.9563} & 1.7246 & \textbf{1.6556} \\
    \bottomrule
    \end{tabular}%
  \label{tbl:syn_small}
\end{table}%

Results on the inference accuracy of two algorithms on selected Covid-19 test dataset with small counts are summarized in Table \ref{tbl:covid_small} where $RG$ represents the range of counts, $PCT_0$ represents the percentage of zero counts, and $PCT_{1,5}$ represents the percentage of counts in [1, 5]. We can see that AGS can outperform NUTS in $R^2$ and RMSE when there are no zero counts (subset1 and subset3) in the covid test data. However, when zero counts are included (subset2 and the whole set), NUTS performs better than AGS. Comparing the performance of both algorithms on subset 2, we can see that even when all the counts are small but positive, AGS can still outperform NUTS in inference accuracy by a small margin.

We also compare the inference accuracy of both algorithms using synthetic datasets with small counts (SS1 - SS10) (Table \ref{tbl:syn_small}). It can be observed that for most synthetic datasets  with a relatively large percentage of small counts (SS1-SS3 and SS5-SS8), NUTS outperforms AGS in terms of $R^2$ and RMSE. If the percentage of small counts is not very high (SS9 and SS10), then AGS outperforms NUTS, even when there are zero counts (SS10). 

Based on the performance comparison using synthetic datasets with small counts and Covid-19 test datasets with small counts, we can conclude that when there is a large percentage of small counts, particularly zero counts, NUTS tends to outperform AGS. The specific percentage of small counts in a dataset that leads to a better performance of NUTS than AGS varies with the dataset. That is, depending on the particular dataset, AGS may still outperform NUTS when there is a large percentage of small counts.


\section{Conclusions} \label{sec:conc}

This research proposes a scalable approximate Gibbs sampling algorithm for the HBPRM for grouped count data. Our algorithm builds on the approximation of data-likelihood with Gaussian distribution such that the conditional posterior for coefficients have a close-form solution. Empirical examples using synthetic and real datasets demonstrate that the proposed algorithm outperforms the state-of-the-art sampling algorithm, NUTS, in inference efficiency. The improvement in efficiency is greater for larger datasets, suggesting improved scalability. Due in part to the Gibbs updates, the AGS trades off greater accuracy for slower mixing Markov chains, leading to a much lower effective sample size and therefore lower sampling efficiency. However, when sampling time is of great concern to model users (e.g. predicting incidents and demands to allocate resources during a disaster), AGS would be the only feasible option. As the approximation quality improves with larger counts, our algorithm works better for count datasets in which the counts are large. When a large portion of counts in a dataset are zero or very small counts, then NUTS tend to outperform AGS in inference accuracy. Therefore, when there are zero counts and inference accuracy is critical, NUTS is recommended over AGS.

It is worth noting that the approximate conditional distributions of the parameters in the HBPRMs for grouped count data may not be compatible with each other, i.e., there may not exist an implicit joint posterior distribution~\cite{gelman2004parameterization, alquier2016noisy} after applying the approximation. However, despite potentially incompatible conditional distributions, the use of such approximate MCMC samplers is suggested due to the computational efficiency and analytical convenience~\cite{gelman2004parameterization,johndrow2015optimal}, especially when the efficiency improvement outweighs the bias introduced by approximation~\cite{alquier2016noisy}. 

Future work can explore scalable inference in hierarchical Bayesian models for data with excessive zeros~\cite{gholiabad2021multilevel,liu2017simulating,brown2015zero} as the Poisson regression model is not appropriate for zero-inflated count data.

\section*{Acknowledgment} \label{sec:acknowledgement}
\noindent
This research was partially funded by the National Science Foundation (grant no. 1635717). We would also like to thank Prof. Qian Qin from the University of Minnesota for his helpful comments on an earlier draft of this paper.


\bibliographystyle{tfnlm}
\bibliography{main_refs.bib}

\section{Appendices}

\appendix
\section{Derivation of the approximate conditional posterior} \label{app:derivation}
\noindent 
Derivation of the approximate posterior distribution in the posterior distribution is presented below. Terms that do not impact $     w_{jk}$ are regarded as a constant, i.e. $C_i$ ($i=1,\dots,5$) in the following equations.

The conditional posterior of regression coefficient $     w_{jk}$ can be written as
\begin{equation}
    p\left( {{     w _{jk}}| - } \right) \propto \exp\left\{ {\frac{{{{\left( {{     w _{jk}} - {\mu _{{k}}}} \right)}^2}}}{{ - 2\sigma _{{k}}^2}} + \sum\limits_{i = 1}^{{n_j}} {\frac{{{{\left[ {{     x_{ijk}}{     w _{jk}} + \sum\limits_{h = 1,h \ne k}^K {{     x_{ijh}}{w _{jh}}}  - {\psi _0}\left( {{     y_{ij}}} \right)} \right]}^2}}}{{ - 2{\psi _1}\left( {{     y_{ij}}} \right)}}} } \right\}.  \label{eq:approx_norm_orig}
\end{equation}

\noindent Let $A$ be the exponent in Eq. \eqref{eq:approx_norm_orig} and expand the square terms, then we have

\begin{align}
A &= \frac{{{{\left( {{     w_{jk}} - {\mu _k}} \right)}^2}}}{{ - 2\sigma _k^2}} + \sum\limits_{i = 1}^{{n_j}} {\frac{{{{\left( {{     w_{jk}}{     x_{ijk}} + \sum\limits_{h = 1,h \ne k}^K {{     w_{jh}}{     x_{ijh}}}  - {\psi _0}\left( {{     y_{ij}}} \right)} \right)}^2}}}{{ - 2{\psi _1}\left( {{     y_{ij}}} \right)}}} \\
 &= \frac{{     w_{jk}^2 - 2{\mu _k}{     w_{jk}} + {C_1}}}{{ - 2\sigma _k^2}} + \sum\limits_{i = 1}^{{n_j}} {\frac{{     x_{ijk}^2     w_{jk}^2 + 2\left( {\sum\limits_{h = 1,h \ne k}^K {{     w_{jh}}{     x_{ijh}}}  - {\psi _0}\left( {{     y_{ij}}} \right)} \right){     x_{ijk}}{     w_{jk}} + {C_2}}}{{ - 2{\psi _1}\left( {{     y_{ij}}} \right)}}} \\
& =  - \frac{1}{2}\left( {\frac{1}{{\sigma _k^2}} + \sum\limits_{i = 1}^{{n_j}} {\frac{{     x_{ijk}^2}}{{{\psi _1}\left( {{     y_{ij}}} \right)}}} } \right)     w_{jk}^2 + \left( {\frac{{{\mu _k}}}{{\sigma _k^2}} + \sum\limits_{i = 1}^{{n_j}} {\frac{{\left( {\sum\limits_{h = 1,h \ne k}^K {{     w_{jh}}{     x_{ijh}}}  - {\psi _0}\left( {{     y_{ij}}} \right)} \right){     x_{ijk}}}}{{ - {\psi _1}\left( {{     y_{ij}}} \right)}}} } \right){     w_{jk}} + {C_3}.
\end{align}

\noindent Dividing the numerator and denominator by the coefficient of the quadratic term, we get

\begin{align}
A & \propto \frac{{     w_{jk}^2 + \left( {\frac{{\frac{{{\mu _k}}}{{\sigma _k^2}} + \sum\limits_{i = 1}^{{n_j}} {\frac{{\left( {\sum\limits_{h = 1,h \ne k}^K {{     w_{jh}}{     x_{ijh}}}  - {\psi _0}\left( {{     y_{ij}}} \right)} \right){     x_{ijk}}}}{{ - {\psi _1}\left( {{     y_{ij}}} \right)}}} }}{{ - \frac{1}{2}\left( {\frac{1}{{\sigma _k^2}} + \sum\limits_{i = 1}^{{n_j}} {\frac{{     x_{ijk}^2}}{{{\psi _1}\left( {{     y_{ij}}} \right)}}} } \right)}}} \right){     w_{jk}}}}{{\frac{1}{{ - \frac{1}{2}\left( {\frac{1}{{\sigma _k^2}} + \sum\limits_{i = 1}^{{n_j}} {\frac{{     x_{ijk}^2}}{{{\psi _1}\left( {{     y_{ij}}} \right)}}} } \right)}}}} + {C_4} \\
& \propto \frac{{{{\left( {{     w_{jk}} - \frac{{\frac{{{\mu _k}}}{{\sigma _k^2}} - \sum\limits_{i = 1}^{{n_j}} {\frac{{\left( {\sum\limits_{h = 1,h \ne k}^K {{     w_{jh}}{     x_{ijh}}}  - {\psi _0}\left( {{     y_{ij}}} \right)} \right){     x_{ijk}}}}{{{\psi _1}\left( {{     y_{ij}}} \right)}}} }}{{\frac{1}{{\sigma _k^2}} + \sum\limits_{i = 1}^{{n_j}} {\frac{{     x_{ijk}^2}}{{{\psi _1}\left( {{     y_{ij}}} \right)}}} }}} \right)}^2}}}{{ - 2 \cdot \frac{1}{{\frac{1}{{\sigma _k^2}} + \sum\limits_{i = 1}^{{n_j}} {\frac{{     x_{ijk}^2}}{{{\psi _1}\left( {{     y_{ij}}} \right)}}} }}}} + {C_5}.
\end{align}

\noindent Therefore, we obtain the mean and variance of the  the approximate Gaussian posterior 

\begin{align}
\widehat{\mu}_{{     {{k}}}} =& \, {\frac{{\frac{{{\mu _k}}}{{\sigma _k^2}} - \sum\limits_{i = 1}^{{n_j}} {\frac{{\left( {\sum\limits_{h = 1,h \ne k}^K {{     w_{jh}}{     x_{ijh}}}  - {\psi _0}\left( {{     y_{ij}}} \right)} \right){     x_{ijk}}}}{{{\psi _1}\left( {{     y_{ij}}} \right)}}} }}{{\frac{1}{{\sigma _k^2}} + \sum\limits_{i = 1}^{{n_j}} {\frac{{     x_{ijk}^2}}{{{\psi _1}\left( {{     y_{ij}}} \right)}}} }}} = \frac{{{\mu _k} + \sigma _k^2\sum\limits_{i = 1}^{{n_j}} {\frac{{{     x_{ijk}}}}{{{\psi _1}\left( {{     y_{ij}}} \right)}}\left( {{\psi _0}\left( {{     y_{ij}}} \right) - \sum\limits_{h = 1,h \ne k}^K {{     w_{jh}}{     x_{ijh}}} } \right)} }}{{\sigma _k^2\sum\limits_{i = 1}^{{n_j}} {\frac{{     x_{ijk}^2}}{{{\psi _1}\left( {{     y_{ij}}} \right)}} + 1} }}, \\ \nonumber
&\,\, \forall j=1,\dots,J,
\end{align}
\begin{align}
{\widehat{\sigma}_{{     {{k}}}}}^2 =& \frac{1}{{\frac{1}{{\sigma _k^2}} + \sum\limits_{i = 1}^{{n_j}} {\frac{{     x_{ijk}^2}}{{{\psi _1}\left( {{     y_{ij}}} \right)}}} }} = \frac{{\sigma _k^2}}{{\sigma _k^2\sum\limits_{i = 1}^{{n_j}} {\frac{{     x_{ijk}^2}}{{{\psi _1}\left( {{     y_{ij}}} \right)}} + 1} }}, \, \forall j=1,\dots,J.
\end{align}

\section{Metrics used for comparing samplers} \label{app:metrics}
\noindent
$\bullet$ \textit{\textbf{Effective sample size}} \textbf{($\hat{n}_{\mathrm{eff}}$)}
\vspace{4pt}

\noindent
For each scalar estimand $\psi$, the simulations/samples are labeled as $\psi_{ij}\;(i=1,\dots,n;j=1,\dots,m)$ where $n$ is the number of samples in each chain (sequence) and $m$ is the number of chains. The effective sample size is calculated according to Ref.~\cite[Chapter 11]{gelman2013bayesian}
\begin{equation}
\hat{n}_{\mathrm{eff}}=\frac{m n}{1+2 \sum_{t=1}^{T} \widehat{\rho}_{t}},
\end{equation}
where the estimated auto-correlations $\widehat{\rho}_{t}$ are computed as
\begin{equation}\label{eq:rho_t}
    \widehat{\rho}_t = 1 - \frac{{{V_t}}}{{2\,{\widehat {\rm var}}^ + }}
\end{equation}
\noindent
and $T$ is the first odd positive integer for which $\widehat{\rho}_{T+1}+\widehat{\rho}_{T+2}$ is negative.
In Eq.~\eqref{eq:rho_t}, $V_t$, the variogram at each lag $t$, is given by
\begin{equation}
{V_t} = \frac{1}{{m\left( {n - t} \right)}}{\sum\limits_{j = 1}^m {\sum\limits_{i = t + 1}^n {\left( {{\psi _{i,j}} - {\psi _{i - t,j}}} \right)^2} } },
\end{equation}

\noindent
and ${\widehat {\rm var}}^ +$, the marginal posterior variance of the estimand, is given by



\begin{equation}
{\widehat {\rm var}}^ +  = \frac{{n - 1}}{mn} \sum\limits_{j = 1}^m {s_j^2} + \frac{1}{{m - 1}}{\sum\limits_{j = 1}^m {\left( {\mathop {{\psi _{. j}}}\limits^ -   - \mathop {{\psi _{. .}}}\limits^ -  } \right)} ^2},
\end{equation}

\noindent
where $s_j^2 = \frac{1}{{n - 1}}{\sum\limits_{i = 1}^n {\left( {{\psi _{ij}} - \mathop {{\psi _{. j}}}\limits^ -  } \right)} ^2},
~\mathop {{\psi _{. j}}}\limits^ -   = \frac{1}{n}\sum\limits_{i = 1}^n {{\psi _{ij}}} ,
~\textrm{and}~\mathop {{\psi _{..}}}\limits^ -   = \frac{1}{m}\sum\limits_{j = 1}^m {\mathop {{\psi _{. j}}}\limits^ -  }.$

\vspace{8pt}
\noindent
$\bullet$ \textit{\textbf{$R^2$}}
\vspace{2pt}

\noindent
The $R^2$ of generic predicted values $\hat y_i, i={1,\dots,N}$ of the dependent variables $y_i, i={1,\dots,N}$ is expressed as
\begin{equation} R^2=1-\frac{\sum_{i=1}^{N}({y_i}-\hat{y_i})^2}{\sum_{i=1}^{N}(y_i-\bar{y})^2}, \end{equation}

\noindent
where $\bar{y}$ is the average value of $y_i, i={1,\dots,N}$.

\vspace{8pt}
\noindent
$\bullet$ \textit{\textbf{RMSE}}
\vspace{2pt}

\noindent
The RMSE of predicted values is given by $\textrm{RMSE}=\sqrt{\frac{\sum_{i=1}^{N}\left(\hat{y}_{i}-y_{i}\right)^{2}}{N}}.$


\end{document}